\documentclass[runningheads]{llncs}
\usepackage{graphicx}

\usepackage{tikz}
\usepackage{comment}
\usepackage{amsmath,amssymb} 
\usepackage{bm}
\usepackage{color}
\usepackage{xspace}

\usepackage{booktabs}
\usepackage{multirow}
\usepackage{pifont}
\definecolor{citecolor}{RGB}{34,139,34}
\usepackage[pagebackref=true,breaklinks=true,letterpaper=true,colorlinks,citecolor=citecolor,bookmarks=false]{hyperref}
\usepackage{colortbl}
\usepackage{makecell}
\usepackage{xfrac}
\usepackage{wrapfig}

\usepackage{pifont}
\newcommand{\cmark}{\ding{51}}
\newcommand{\xmark}{\ding{55}}
\usepackage{subfig}

\makeatletter
\DeclareRobustCommand\onedot{\futurelet\@let@token\@onedot}
\def\@onedot{\ifx\@let@token.\else.\null\fi\xspace}

\def\eg{\emph{e.g}\onedot} 
\def\ie{\emph{i.e}\onedot}

\makeatother

\renewcommand{\Sigma}{\mathfrak{S}}

\def\1{\bm{1}}

\newcommand{\test}{\mathcal{D_{\mathrm{test}}}}

\def\vf{{\bm{f}}}

\def\vk{{\bm{k}}}

\def\vq{{\bm{q}}}

\def\mI{{\bm{I}}}

\def\mM{{\bm{M}}}

\DeclareMathAlphabet{\mathsfit}{\encodingdefault}{\sfdefault}{m}{sl}
\SetMathAlphabet{\mathsfit}{bold}{\encodingdefault}{\sfdefault}{bx}{n}

\def\sF{{\mathbb{F}}}

\begin{document}
\pagestyle{headings}
\mainmatter

\definecolor{defaultcolor}{gray}{0.9}
\definecolor{demphcolor}{gray}{.5}
\newcommand{\demph}[1]{\textcolor{demphcolor}{#1}}
\title{CP$^2$: Copy-Paste Contrastive Pretraining for Semantic Segmentation}

\titlerunning{Copy-Paste Contrastive Pretraining}
%
\author{
  Feng Wang\inst{1} \and Huiyu Wang\inst{2} \and Chen Wei\inst{2} \and Alan Yuille\inst{2} \and Wei Shen\inst{3}\thanks{Corresponding author, \email{shenwei1231@gmail.com}}
}
%
\authorrunning{F. Wang et al.}
%
\institute{Department of Automation, Tsinghua University \and
Department of Computer Science, Johns Hopkins University \and MoE Key Lab of Artificial Intelligence, AI Institute, Shanghai Jiao Tong University}
\maketitle

\begin{abstract}
Recent advances in self-supervised contrastive learning yield good image-level representation, which favors classification tasks but usually neglects pixel-level detailed information, leading to unsatisfactory transfer performance to dense prediction tasks such as semantic segmentation.
In this work, we propose a pixel-wise contrastive learning method called CP$^2$ ({\bf C}opy-{\bf P}aste {\bf C}ontrastive {\bf P}retraining), which facilitates both image- and pixel-level representation learning and therefore is more suitable for downstream dense prediction tasks.
In detail, we copy-paste a random crop from an image (the foreground) onto different background images and pretrain a semantic segmentation model with the objective of 1) distinguishing the foreground pixels from the background pixels, and 2) identifying the composed images that share the same foreground.
Experiments show the strong performance of CP$^2$ in downstream semantic segmentation: By finetuning CP$^2$ pretrained models on PASCAL VOC 2012, we obtain 78.6\% mIoU with a ResNet-50 and 79.5\% with a ViT-S.
Code and models are available at \url{https://github.com/wangf3014/CP2}.

\keywords{dense contrastive learning, semantic segmentation}
\end{abstract}

\section{Introduction}
\label{sec:intro}

Learning invariant {\it image-level} representation and transferring to downstream tasks has became a common paradigm in self-supervised contrastive learning. Specifically, the objective of these methods is either to minimize the Euclidean ($\ell_2$) distance~\cite{byol,simsiam} or cross entropy~\cite{swav,dino} between the {\it image-level} features of augmented views of the same image, or to distinguish the positive image feature from a set of negative image features by optimizing an InfoNCE~\cite{oord2018representation} loss~\cite{moco,mocov2,mocov3,simclr,simclrv2,infomin}.

In spite of the success in downstream classification tasks, these contrastive objectives build on the assumption that every pixel in an image belongs to a single label and lack the perception of spatially varying image content. We argue that these {\it classification-oriented} objectives are not ideal for downstream dense prediction tasks such as semantic segmentation where the model should distinguish different semantic labels in an image. For the task of semantic segmentation, current contrastive learning models may easily over-fit to learning the {\it image-level} representation and neglect pixel-level variances.

Moreover, there is an architectural misalignment in the current pretraining finetuning paradigm for downstream semantic segmentation tasks: 1) The semantic segmentation model usually requires a large atrous rate and a small output stride than those in the classification-oriented pretrained backbones~\cite{long2015fully,chen2017deeplabv3}; 2) The finetuning of the well pretrained backbone and the randomly initialized segmentation head can be out of sync, \eg the random head may generate random gradients that poison the pretrained backbone, negatively affecting the performance. These two issues prevent the classification-oriented pretrained backbone from facilitating dense prediction tasks such as semantic segmentation.

\begin{wrapfigure}{r}{0.4\textwidth}
  \centering
  \includegraphics[width=\linewidth]{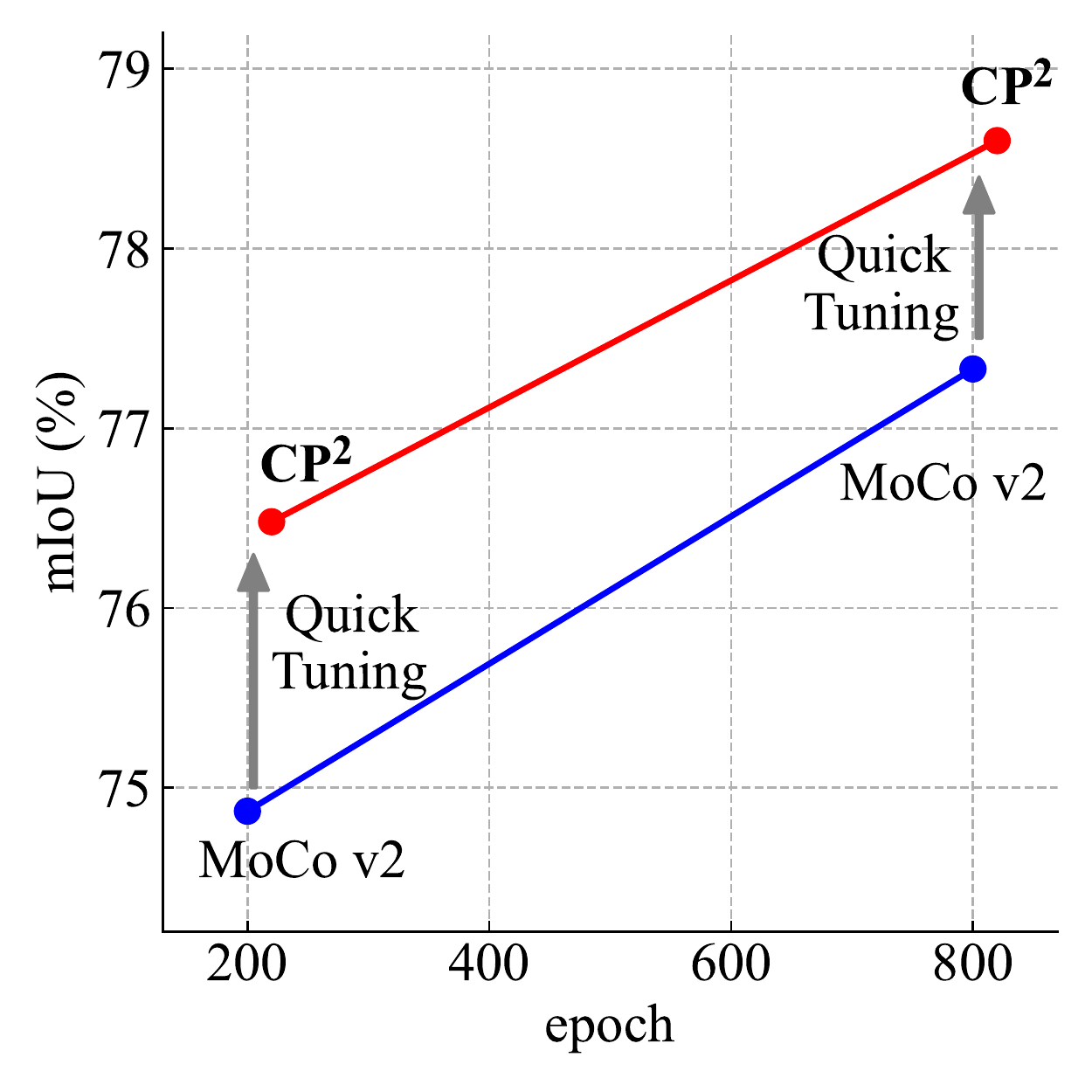}
  \vspace{-20pt}
  \caption{\textbf{Quick Tuning MoCo v2 with CP$^2$}, evaluated by semantic segmentation on PASCAL VOC. A 20-epoch Quick Tuning with CP$^2$ yields large mIoU improvements.}
  \label{fig:perf}
\end{wrapfigure}

In this paper, we propose a novel self-supervised pretraining method designed for downstream semantic segmentation, named {\bf C}opy-{\bf P}aste {\bf C}ontrastive {\bf P}retraining (CP$^2$). Specifically, we pretrain a semantic segmentation model with Copy-Pasted input images which are composed by cropping random crops from a foreground image and pasting them onto different background images. Examples of the composed images are shown in Figures~\ref{fig:pipeline}. Aside from the image-wise contrastive loss for learning instance discrimination~\cite{amdim,wu2018unsupervised,ye2019unsupervised,moco}, we introduce a pixel-wise contrastive loss to enhance dense prediction. The segmentation model is trained by maximizing cosine similarity between the foreground pixels while minimizing cosine similarity between the foreground and background pixels. Overall, CP$^2$ yields pixel specific dense representation and has two key advantages for downstream segmentation: 1) CP$^2$ pretrains both backbone and segmentation head, addressing the issue of architectural misalignment; 2) CP$^2$ pretrains the model with a dense prediction objective, building up the model's perception of spatially varying information in an image.

Furthermore, we find that a considerably short period of CP$^2$ training is able to adapt pretrained classification-oriented models quickly to the semantic segmentation task and therefore yields better downstream performance.
In particular, we first initialize the backbone with the weights of a pretrained classification-oriented model (\eg a ResNet-50~\cite{resnet} pretrained by MoCo v2~\cite{mocov2}), attach a randomly initialized segmentation head, and then tune the entire segmentation model by CP$^2$ for additional 20 epochs. As a result, the performance of the entire segmentation model on downstream semantic segmentation is significantly improved, \eg +{\bf 1.6}\% mIoU on PASCAL VOC 2012~\cite{pascal} dataset.
We denote this training protocol as {\bf Quick Tuning}, as it is efficient and practical for transfer learning from image-level instance discrimination to pixel-level dense prediction.

For technical details, we mostly follow MoCo v2~\cite{mocov2}, including its architecture, data augmentation, and the instance contrastive loss, in order to fully isolate the effectiveness of our newly introduced copy-paste mechanism and dense contrastive loss, and therefore MoCo v2~\cite{mocov2} serves as a direct baseline to CP$^2$. In the empirical evaluations of semantic segmentation, the CP$^2$ 200-epoch model achieves 77.6\% mIoU on PASCAL VOC 2012~\cite{pascal}, outperforming the MoCo v2~\cite{mocov2} 200-epoch model by +{\bf 2.7}\% mIoU. Also, as illustrated in Figure~\ref{fig:perf}, the Quick Tuning protocol for CP$^2$ yields +{\bf 1.5}\% and +{\bf 1.4}\% mIoU improvements over the MoCo v2 200-epoch and 800-epoch model respectively. The improvement also generalizes to other segmentation datasets and vision transformers.

\section{Related Work} \label{sec:relat}
\textbf{Self-supervised learning and pretext tasks.} Self-supervised learning for visual understanding leverages the intrinsic properties of images as the supervisory information for training, for which the capability of visual representation heavily depends on the formulation of pretext tasks. Prior to the recent popularity of instance discrimination~\cite{amdim,wu2018unsupervised,ye2019unsupervised,moco}, people have explored numerous pretext tasks, including image denoising and reconstruction~\cite{pathak2016context,zhang2017split,belghazi2019learning}, adversarial learning~\cite{bigan,bigbigan,dumoulin2016adversarially}, and heuristic tasks such as image colorization~\cite{zhang2016colorful}, jigsaw puzzle~\cite{jigsaw,iterjigsaw}, context and rotation prediction~\cite{pos,rotnet}, and deep clustering~\cite{deepcluster}.

The emergence of contrastive learning, or more specifically, the scheme of instance discrimination~\cite{amdim,wu2018unsupervised,ye2019unsupervised,moco} has made a break-through in unsupervised learning, as MoCo~\cite{moco} achieves superior transfer performance than supervised training in a wide range of downstream tasks. Inspired by this success, many follow-up works conduct deeper explorations in self-supervised contrastive learning and put forward different optimization objectives~\cite{byol,infomin,wei2020co2,swav,dino,zhou2022ibot}, model architectures~\cite{simsiam}, and training strategies~\cite{simclr,mocov2,mocov3}.

\textbf{Dense contrastive learning.} To obtain better adaptation in dense prediction tasks, a recent work~\cite{o2020unsupervised} extends the image-level contrastive loss into a pixel-level. Despite the extension of contrastive loss helps the model learn finer grained features, it is not able to establish the model's perception of spatially varying information, and therefore the model has to be re-purposed in downstream finetuning. More recent works try to enhance the model's understanding of positional information in images by encouraging the consistency of pixel-level representations~\cite{xie2021propagate}, or by employing heuristic masks~\cite{felzenszwalb2004efficient,arbelaez2014multiscale} and applying a patch-wise contrastive loss~\cite{henaff2021efficient}.

\textbf{Copy-paste for contrastive learning.} Copy-paste, \ie, copying crops of one image and pasting them onto another image, once serves as a data augmentation method in {\it supervised} instance segmentation and semantic segmentation~\cite{ghiasi2021simple} for its simplicity and significant effect in enriching images' positional and semantic information. Similarly, by mixing images~\cite{mixup,augmix} or image crops~\cite{cutmix} as data augmentation, the {\it supervised} models also attain considerable performance improvements in various tasks. The use of copy-paste is also reported in recent works of self-supervised object detection~\cite{yang2021instance,henaff2021efficient}. Inspired by the success of copy-paste, we utilize this approach in our dense contrastive learning method as the self-supervisory information.

\section{Method}
\label{sec:method}

In this section, we present our CP$^2$ objective and loss function for learning a pixel-wise dense representation. We also discuss our model architecture and propose a Quick Tuning protocol for efficient training of CP$^2$.

\begin{figure}[t]
  \centering
  \includegraphics[width=0.9\linewidth]{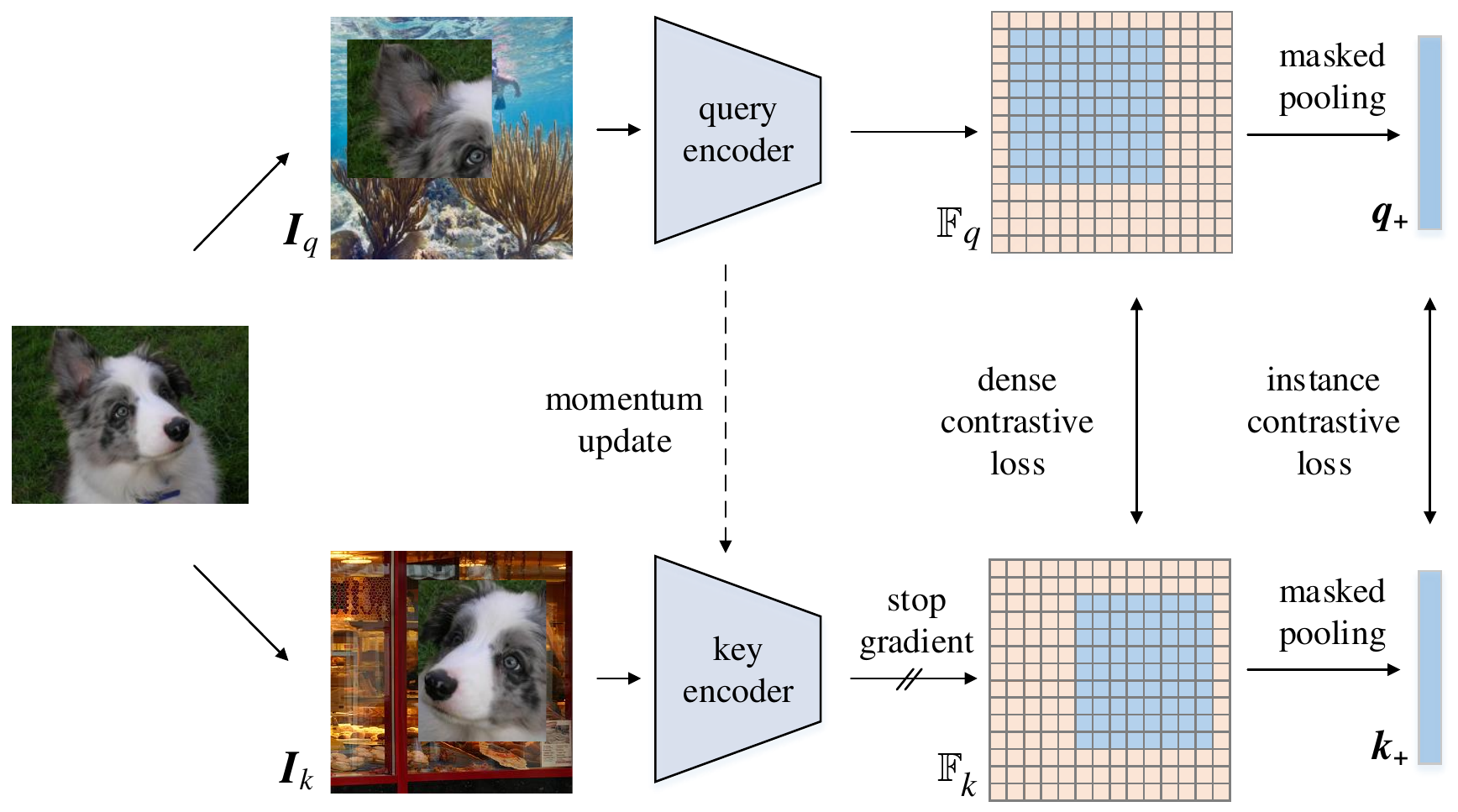}
  \caption{\textbf{Pipeline}. We enrich the spatial information of unannotated images by randomly pasting two crops of foreground images onto different backgrounds. A dense contrastive loss is applied to their encoded feature maps and an instance contrastive loss is applied to the average of the foreground feature vectors (masked pooling). We follow the training architecture of momentum update in MoCo and BYOL.}
  \label{fig:pipeline}
\end{figure}

\subsection{Copy-paste contrastive pretraining}
\label{sec:pretext}
We propose a novel pretraining method called CP$^2$, through which we desire the pretrained model to learn both instance discrimination and dense prediction. To this end, we manually synthesize image compositions by pasting foreground crops onto backgrounds. Specifically, as illustrated in Figure~\ref{fig:pipeline}, we generate two random crops from the foreground image and then overlay them onto two different background images. The objective of CP$^2$ is to 1) discriminate the foreground from background within each composed image and 2) identify the composed images with the same foreground from negative samples.

\textbf{Copy-paste.}
Given an original foreground image $\mI^{fore}$, we first generate two different views of it $\mI_q^{fore},\ \mI_{k}^{fore}\in\mathbb{R}^{224\times224\times3}$ by data augmentation, one being query and the other being the positive key. The augmentation strategy follows SimCLR~\cite{simclr} and MoCo v2~\cite{mocov2}, \ie, the image is first randomly resized and cropped to 224$\times$224 resolutions followed by color jittering, gray scale, Gaussian blurring and horizontal flipping. Next, we generate one view for each of two random background images using the same augmentation, denoted as $\mI_q^{back},\ \mI_{k}^{back}\in\mathbb{R}^{224\times224\times3}$. We compose the image pairs by binary foreground-background masks $\mM_q,\mM_{k}\in\{0,1\}^{224\times224}$, in which each element $m=1$ denotes a foreground pixel and $m=0$ denotes a background pixel. Formally, the composed images are generated by
\begin{equation}
\begin{split}
    \mI_q=\mI_q^{fore} \odot \mM_q + \mI_q^{back} \odot (1 - \mM_q)\,,\\
    \mI_{k}=\mI_{k}^{fore} \odot \mM_{k} + \mI_{k}^{back} \odot (1 - \mM_{k})\,,
\end{split}
\end{equation}
where $\odot$ denotes element-wise product. Now we get two composed images $\mI_q$ and $\mI_k$ who share the foreground source image but have different backgrounds.

\begin{wrapfigure}{r}{0.45\textwidth}
    \centering
    \includegraphics[width=\linewidth]{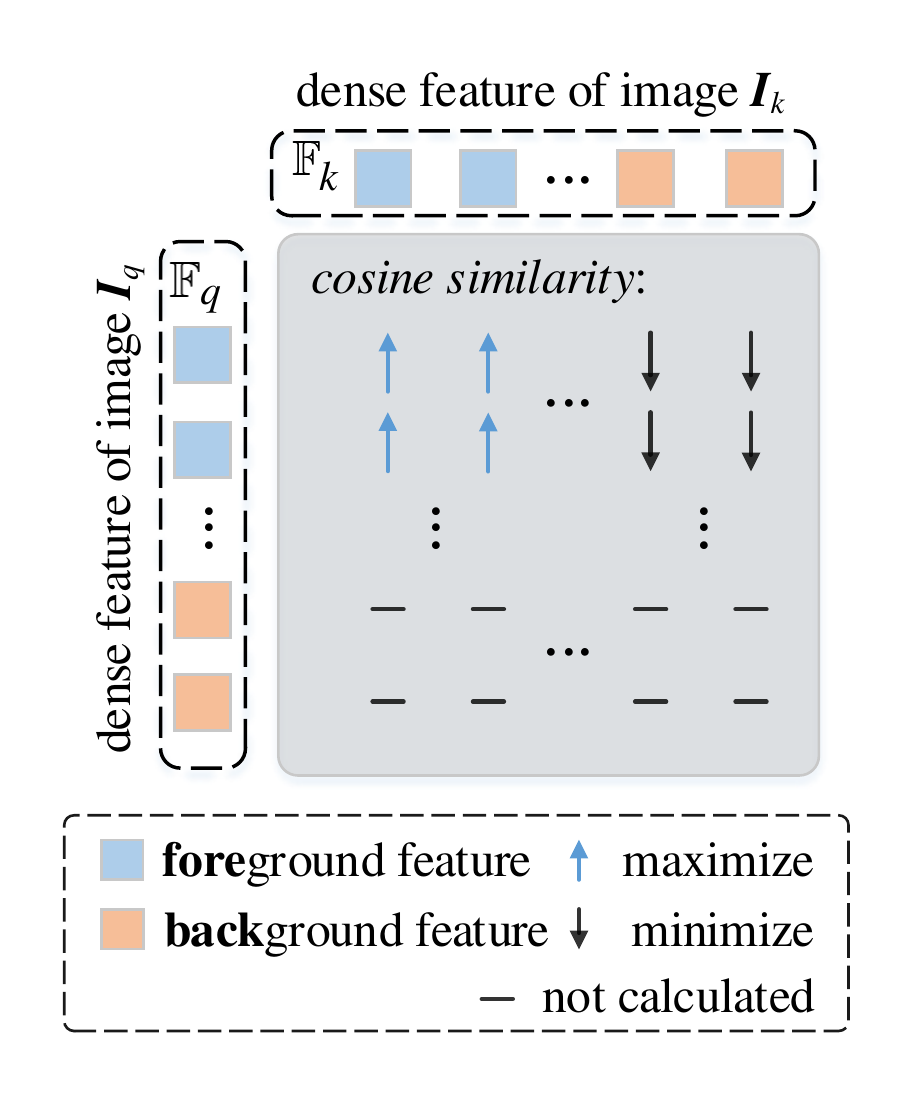}
    \caption{\textbf{Dense contrastive loss} that maximizes the similarity of each foreground pair while minimizes that of each foreground-background pair.}
    \label{fig:dense}
\end{wrapfigure}

\textbf{Contrastive objectives.}
The composed images are then processed by a semantic segmentation model which we detail in Section~\ref{sec:arc}. Given the input $\mI_q$, the output of the segmentation model is a set of $r\times r$ features $\sF_q=\{\vf^i_q \in \mathbb{R}^C | i = 1, 2, \ldots, r^2 \}$, where $C$ is the number of output channels and $r$ is the feature map resolution. For a $224\times 224$ input image, $r=14$ when the output stride $s=16$. Among all the output features $\vf_q \in \sF_q$, we denote the foreground features, \ie, the features that correspond to foreground pixels as $\vf^+_q \in \sF^+_q \subset \sF_q$, where $\sF^+_q$ is the foreground feature subset. Similarly, we have all the features $\vf_{k} \in \sF_{k}$ for the input image $\mI_{k}$, among which the foreground features are denoted as $\vf^+_{k} \in \sF^+_{k} \subset \sF_{k}$.

We use two loss terms, one \textit{dense} contrastive loss and one \textit{instance} contrastive loss. The contrastive loss $\mathcal{L}_{dense}$ learns local and fine-grained features by distinguishing between foreground and background features, helping with downstream semantic segmentation tasks, while the instance contrastive loss aims to keep the global, instance-level representation.

In dense contrastive loss, we desire all the foreground features $\forall \vf^+_q \in \sF^+_q$ of image $\mI_q$ to be similar to all the foreground features $\forall \vf^+_k \in \sF^+_k$ of image $\mI_k$, and be dissimilar to the background features $\sF^-_k = \sF_k \setminus \sF^+_k$ of image $\mI_k$. Formally, for each foreground feature $\forall \vf^+_q \in \sF^+_q$ and $\forall \vf^+_k \in \sF^+_k$, the dense contrastive loss is obtained by
\begin{equation}
    \mathcal{L}_{dense}=-\frac{1}{|\sF^+_{q}||\sF^+_{k}|}
    \sum_{\forall \vf_q^+\in \sF_q^+, \forall \vf_k^+\in \sF_k^+}\log{
    \frac{\exp{(\vf_q^+\cdot \vf_k^+/\tau_{dense})}}{
    \sum_{\forall \vf_k \in \sF_k}\exp{(\vf_q^+\cdot \vf_k/\tau_{dense})}
    }}\,,
\end{equation}
where $\tau_{dense}$ is a temperature coefficient. This dense contrastive loss is also illustrated in Figure~\ref{fig:dense}. Following supervised contrastive learning methods~\cite{supervisedcl,zhao2021contrastive}, we put the summation outside the log.

Besides the dense contrastive loss, we keep the instance contrastive loss that aims to learn the global, instance-level representation. We mostly follow the practice of MoCo~\cite{moco,mocov2}, where given the query image, the model is required to distinguish the positive key from a memory bank of negative keys. But in our case, we use the composed image $\mI_q$ as the query image, and the composed image $\mI_k$ as the positive key image that shares the foreground with image $\mI_q$. In addition, instead of using the global average pooling feature as the representation in MoCo, we use the normalized masked averaging of only the foreground features as illustrated in Figure~\ref{fig:pipeline}. Formally, the instance contrastive loss is computed as
\begin{equation}
    \mathcal{L}_{ins}=-\log\frac{
    \exp(\vq_+\cdot\vk_+/\tau_{ins})
    }{
    \exp(\vq_+\cdot\vk_+/\tau_{ins}) + \sum_{n=1}^N{\exp(\vq_+\cdot\vk_n/\tau_{ins})}}\,,
\end{equation}
where $\vq_+$, $\vk_+$ are normalized masked averaging of $\sF^+_q$ and $\sF^+_k$:
\begin{equation}
    \vq_+= \frac{\sum\nolimits_{\forall\vf_q^+\in \sF_q^+}{\vf_q^+}}{||\sum\nolimits_{\forall\vf_q^+\in \sF_q^+}{\vf_q^+}||_2}, \ \vk_+=\frac{\sum\nolimits_{\forall\vf_k^+\in \sF_k^+}{\vf_k^+}}{||\sum\nolimits_{\forall\vf_k^+\in \sF_k^+}{\vf_k^+}||_2}\,.
\end{equation}
$\vk_n$ denotes the representations of negative samples from a memory bank~\cite{moco,wu2018unsupervised} of $N$ vectors, and $\tau_{ins}$ is a temperature coefficient.

The total loss $\mathcal{L}$ is simply a linear combination of the dense and the instance contrastive loss
\begin{equation}
    \mathcal{L}=\mathcal{L}_{ins}+\alpha\mathcal{L}_{dense},
\end{equation}
where $\alpha$ is a trade-off coefficient for the two losses.

\subsection{Model architecture}
\label{sec:arc}
Next, we discuss in detail our CP$^2$ model architecture that consists of a backbone and a segmentation head for both pretraining and finetuning. Different from existing contrastive learning frameworks~\cite{moco,mocov2} that pretrain only the backbone, CP$^2$ enables the pretraining of both the backbone and the segmentation head, almost the same architecture as the one used for downstream segmentation tasks. In this way, CP$^2$ prevents the finetuning misalignment issue (Section~\ref{sec:intro}), \ie, finetuning the downstream models with a well-pretrained backbone and a randomly initialized head. This misalignment can require careful hyper-parameter tuning (\eg, a larger learning rate on the head) and result in degradation of the transferring performance, especially when a heavy randomly initialized head is used. Therefore, CP$^2$ is able to achieve better performance for segmentation and also enables the usage of stronger segmentation heads.

In particular, we study two families of backbones, CNNs~\cite{resnet} and vision transformers~\cite{vit}. For CNN backbones, we use the original ResNet-50~\cite{resnet} with a 7$\times$7 convolution as the first layer, instead of an inception stem~\cite{szegedy2017inception} commonly used in segmentation tasks~\cite{chen2018deeplabv2,chen2017deeplabv3,deeplabv3plus2018}. This setting ensures fair comparisons with previous self-supervised learning methods. In order to adapt the ResNet backbone to segmentation, we follow common segmentation settings~\cite{chen2018deeplabv2,chen2017deeplabv3,moco} and use atrous rate 2 and stride 1 for the 3×3 convolutions in the last stage. For vision transformer backbones, we choose ViT-S~\cite{vit} with 16$\times$16 patch size, which has a similar number of parameters as ResNet-50. Note that both of our ResNet-50 and ViT-S have an output stride $s=16$ which makes our backbones compatible with most existing segmentation heads.

Given the backbone output features with an output stride $s=16$, we study two types of segmentation heads. By default, we employ the common DeepLab v3~\cite{chen2017deeplabv3} segmentation head (\ie ASPP head with image pooling), as it is able to extract multi-scale spatial features and yield very competitive results. In addition to the DeepLab v3 ASPP head, we also study the lightweight FCN head~\cite{long2015fully} usually adopted for evaluation of self-supervised learning methods.

On top of the backbone and segmentation head that are trained for both pretraining and finetuning, we make as little change as possible. Specifically, for CP$^2$ pretraining, we add two 1$\times$1 convolution layers to the segmentation head output, projecting the pixel-wise dense features to a 128-dimensional latent space (\ie, $C=128$). The latent features at each pixel are then $\ell_2$ normalized individually. Our dense projection design is analogous to the 2-layer MLP design in common contrastive learning frameworks~\cite{mocov2} followed by an $\ell_2$ normalization. After the CP$^2$ training converges, we simply replace the 2-layer convolution projection by a segmentation output convolution that projects the segmentation head feature to the number of output classes, similar to the typical design in image-wise contrastive frameworks~\cite{moco,simclr}. Following MoCo~\cite{moco}, we momentum update the key encoder consisting of both the backbone and the segmentation head by the weights in the query encoder.

\subsection{Quick Tuning}
\label{sec:quick}
In order to train our CP$^2$ models quickly within a manageable computational budget, we propose a new training protocol called Quick Tuning that initializes our backbone with existing backbone checkpoints available online. These backbones typically have been trained by image-wise contrastive loss with extremely long schedules (\eg 800 epochs~\cite{mocov2} or 1000 epochs~\cite{simclr}). On top of these existing checkpoints that encode good image-level semantic representations, we apply our CP$^2$ training for just a few epochs (\eg, 20 epochs) in order to finetune the representation still on ImageNet without human labels but for semantic segmentation. Specifically, we attach a randomly initialized segmentation head on top of the pretrained backbone with proper atrous rates and train this entire segmentation model with our CP$^2$ loss function. Finally, the learned segmentation model on ImageNet without using any label is further finetuned on various downstream segmentation datasets for evaluation of the learned representations.

Quick Tuning enables efficient and practical training for self-supervised contrastive learning, as it exploits the heavily-pretrained self-supervised backbones and let them quickly adapt to the desired objective or downstream tasks. According to our empirical evaluations, 20 epochs of Quick Tuning is sufficient to yield significant improvements on various datasets (for example, the finetuning mIoU on PASCAL VOC 2012 is improved by 1.6\% after a 20-epoch Quick Tuning). This is particularly helpful for pretraining segmentation models efficiently, because segmentation models are usually much heavier than the backbone in terms of computational cost due to the atrous convolutions in the backbone and the ASPP module. In this case, Quick Tuning saves a large amount of computational resources by demonstrating significant improvements with a short period of segmentation model self-supervised pretraining.

\section{Experiments}
\label{sec:expri}

\subsection{Experimental setup}
Our MoCo v2 implementation follows the official open source code~\cite{mocov2}, and our semantic segmentation implementation uses the MMSegmentation~\cite{mmseg2020} library.

\textbf{Datasets.} We pretrain CP$^2$ and the baselines on ImageNet~\cite{imagenet} ($\sim$1.28 million training images) and finetune on semantic segmentation tasks of PASCAL VOC~\cite{pascal}, Cityscapes~\cite{cordts2016cityscapes}, and ADE20k~\cite{zhou2019semantic}. For PASCAL VOC, we train on the augmented training set~\cite{vocaug} with 10582 images and evaluate on VOC2012 validation set. For Cityscapes, we train on the ``train-fine'' set with 2975 images and evaluate on its validation set. For ADE20k, we train on the training set with 20210 images and evaluate on the validation set.

\textbf{Segmentation and projection heads.} Our DeepLab v3 ASPP head follows the default setting in MMSegmentation which uses 512 output channels for both the atrous convolutions and the output projection. Our CP$^2$ projection head consists of two layers of 512-channel 1$\times$1 convolutions, ReLU, and a $C=128$ channel 1$\times$1 convolution. For the FCN head, we follow the settings in prior works~\cite{moco,henaff2021efficient} for fair comparison, \ie, two layers of 256-channel 3$\times$3 convolutions with atrous rate$=$6 followed by BN and ReLU. The CP$^2$ projection for the FCN-based model consists of two layers of 256-channel 1$\times$1 convolutions, ReLU, and a $C=128$ channel 1$\times$1 convolution.

\textbf{Baselines.} We compare CP$^2$ with the self-supervised contrastive learning methods with classification-oriented~\cite{moco,mocov2,simclr,infomin,byol}, detection-oriented~\cite{yang2021instance,henaff2021efficient}, and dense prediction~\cite{xie2021propagate} objectives. All the pretrained ResNet-50 models of are downloaded from their official implementations.
For InsLoc~\cite{yang2021instance}, we use the backbone of its ResNet50-FPN model which has been pretrained for 400 epochs. For DetCon~\cite{henaff2021efficient}, we use the model pretrained 1000 epochs with DetCon-B manner for the most competitive baseline. The pretrained ViT-S model of MoCo v2 is borrowed from DINO~\cite{dino}. Moreover, to compare with supervised methods, we load a pretrained ResNet-50 model in torchvision official model zoo, which has a top-1 accuracy of 76.13\% on ImageNet validation set~\cite{imagenet}.

\textbf{Hyper-parameters.} For ResNet-backed models, we pretrain by SGD optimizer with 0.03 learning rate, 0.9 momentum, 0.0001 weight decay, and a mini-batch size of 256 on ImageNet. We finetune them by SGD with 0.9 momuntum, 0.0005 weight decay, and 0.003, 0.01, 0.01 learning rate for PASCAL, Cityscapes, and ADE20k, respectively. For ViT-backed models, we also pretrain with a mini-batch size of 256 on ImageNet but apply an AdamW~\cite{adamw} optimizer with $\beta_1=0.9$,\ $\beta_2=0.999$, 0.00005 learning rate, 0.01 weight decay for both pretraining and finetuning. We pretrain and finetune with 4 GPUs. We find that for CP$^2$ pretrained models, the use of weight decay in finetuning stage usually leads to $\sim$0.2\% mIoU decrease. This is possibly because for the baseline methods, the segmentation head is randomly initialized in finetuning and relies on weight decay for better generalization. However, as CP$^2$ pretrains both the backbone and segmentation head with a proper weight decay, the ewights of segmentation head have been decayed into a lower scale and do not require further decaying during finetuning. Thus, in the finetuning stage, we use weight decay for those baseline models with random segmentation head and turn off weight decay for CP$^2$ models. We adopt a memory bank of $N=65536$, $C=128$ dimensional vectors, with $\boldsymbol{k_+}$, the normalized masked average representation of image $I_k$ in the current mini-batch enqueued and the oldest vectors dequeued. For instance contrastive loss $\mathcal{L}_{ins}$, we set the temperature $\tau_{ins}=0.2$ in accordance with MoCo v2~\cite{mocov2}. We assign a weight of $\alpha=0.2$ and set the temperature $\tau_{dense}=1$ for $\mathcal{L}_{dense}$, according to grid search. For PASCAL VOC, we use crop size 512$\times$512 and train with batch size 16 for 40k iterations. For Cityscapes, we use crop size 512$\times$1024 and train with batch size 8 for 60k iterations. For ADE20k, we use crop size 512$\times$512 and train with batch size 16 for 80k iterations.

\subsection{Main results}

MoCo v2~\cite{mocov2} is a direct baseline to our method as we follow its model architecture, contrastive formulation, and the technical setups. For ease of reference, we use the following abbreviations to denote MoCo v2 pretrained models:
\begin{quote}
    \begin{itemize}
        \item {\bf r.200}, {\bf r.800}: ResNet-50 pretrained by MoCo v2 for 200, 800 epochs.
        \item {\bf v.300}: ViT-S/16 pretrained by MoCo v2 for 300 epochs.
    \end{itemize}
\end{quote}

\begin{table}[t]
    \caption{\textbf{Evaluation results (mIoU) with DeepLab v3 segmentation head}. QT denotes Quick Tuning with CP$^2$ initialized by a MoCo v2 pre-trained backbone. Our results are marked in \colorbox{gray!20}{gray}. The best results are {\bf bolded}. Epochs that are consumed by the initialization model are \demph{de-emphasized}.  }
    \centering
    \begin{tabular}{m{2.7cm}|m{1.75cm}|m{1.5cm}<{\centering}|m{1.6cm}<{\centering}m{1.75cm}<{\centering}m{1.6cm}<{\centering}}
    method & backbone & epoch & PASCAL & Cityscapes & ADE20k\\
    \Xhline{3\arrayrulewidth}
    supervised & ResNet-50 & - & 76.0 & 76.3 & 39.5\\
    \hline
    MoCo~\cite{moco} & ResNet-50 & 200 & 73.2 & 75.8 & 38.6\\
    SimCLR~\cite{simclr} & ResNet-50 & 1000 & 77.3 & 76.5 & 40.1\\
    BYOL~\cite{byol} & ResNet-50 & 300 & 77.4 & 76.5 & 40.2 \\
    InfoMin~\cite{infomin} & ResNet-50 & 800 & 77.2 & 76.5 & 39.6\\\hline
    InsLoc~\cite{yang2021instance} & ResNet-50 & 400 & 75.6 & 76.3 & 40.3\\
    DetCon~\cite{henaff2021efficient} & ResNet-50 & 1000 & 78.1 & 77.1 & 40.6\\
    PixPro~\cite{xie2021propagate} & ResNet-50 & 400 & 77.5 & 76.6 & 40.3\\\hline
    MoCo v2~\cite{mocov2} & ResNet-50 & 200 & 74.9 & 76.2 & 39.2\\
    \rowcolor{gray!20}
    {\bf CP$^2$} & ResNet-50 & 200 & 77.6 & 77.3 & 40.5\\
    \rowcolor{gray!20}
    {\bf CP$^2$} QT r.200 & ResNet-50 & \demph{200+}20 & 76.5 & 77.2 & 40.7\\
    MoCo v2~\cite{mocov2} & ResNet-50 & 800 & 77.2 & 76.4 & 39.7\\
    \rowcolor{gray!20}
    {\bf CP$^2$} QT r.800 & ResNet-50 & \demph{800+}20 & {\bf 78.6} & {\bf 77.4} & {\bf 41.3}\\
    \Xhline{2\arrayrulewidth}
    MoCo v2~\cite{mocov2} & ViT-S/16 & 300 & 78.8 & 77.2 & 41.3\\
    \rowcolor{gray!20}
    {\bf CP$^2$} QT v.300 & ViT-S/16 & \demph{300+}20 & {\bf 79.5} & {\bf 77.6} & {\bf 42.2}\\
    \end{tabular}
    \label{tab:main1}
\end{table}

\textbf{Results with DeepLab v3 segmentation head.} We first present the evaluation results of DeepLab v3 semantic segmentation models (a backbone attached by an ASPP head with image pooling)~\cite{chen2017deeplabv3}. As summarized in Table~\ref{tab:main1}, CP$^2$ achieves 77.6\% mIoU on PASCAL VOC with 200 epochs pretraining from scratch using a ResNet-50 backbone, which outperforms MoCo v2 by +{\bf 2.7}\%. Also, the Quick Tuning protocol is demonstrated to be effective as it yields +{\bf 1.6}\% mIoU on PASCAL VOC when tuning a 200-epoch MoCo v2 checkpoint for only another 20 epochs with CP$^2$, and +{\bf 1.4}\% mIoU when tuning an 800-epoch MoCo v2 checkpoint. Moreover, by Quick Tuning the 800-epoch MoCo v2 model, CP$^2$ achieves the best performance among all ResNet-50 based methods on three evaluated datasets. Notably, it yields +{\bf 0.5}\% mIoU on PASCAL VOC and +{\bf 0.7}\% mIoU on ADE20k compared with the most competitive DetCon~\cite{henaff2021efficient}, in spite of DetCon's heavier computational cost and longer training schedule. For ViT based models, CP$^2$ also outperforms its MoCo v2 baseline by +{\bf 0.7}\% mIoU on PASCAL, +{\bf 0.4}\% mIoU on Cityscapes, and +{\bf 0.9}\% mIoU on ADE20k when Quick Tuning for another 20 epochs.

\textbf{Results with FCN segmentation head.} Table~\ref{tab:main2} summarizes the evaluation results with the light-weight FCN~\cite{long2015fully} head (two hidden layers of atrous convolutions and a classification layer). Similarly, CP$^2$ achieves the highest mIoU on the three datasets with both ResNet-backed and ViT-backed architectures. In particular, compared to the baseline MoCo v2, CP$^2$ obtains up to +{\bf 1.0}\% mIoU on PASCAL using ResNet-50 and +{\bf 0.9}\% mIoU using ViT-S.

Overall, CP$^2$ yields significant performance improvements in the downstream task of semantic segmentation with both strong (ASPP) and light-weight (FCN) segmentation heads. Aside from demonstrating the effectiveness and robustness of CP$^2$ in terms of different segmentation heads, we further dissect the performance improvements from various factors and components in our ablation study. The more in-depth discussion and results in the ablation study show that our improvements on downstream segmentation tasks do not merely come from pretraining the segmentation head.

\subsection{Ablation study}
\label{sec:abl}

In this section, we first question if CP$^2$ benefits downstream semantic segmentation tasks only because it offers a pretrained segmentation head, or the proposed dense contrastive loss ($\mathcal{L}_{dense}$) also helps? Second, we explore the effect of various types of copy-paste masks, ranging from a simple rectangle mask to masking random patches. Third, we study the effect of the training schedule in Quick Tuning. Fourth, we study the effect of two key hyper-parameters, the loss coefficient ($\alpha$) and temperature ($\tau_{dense}$) of the dense contrastive loss.

\begin{table}[t]
    \caption{\textbf{Evaluation results (mIoU) with FCN head}. QT denotes Quick Tuning with CP$^2$ initialized by a MoCo v2 pre-trained backbone. Our results are marked in \colorbox{gray!20}{gray}. The best results are {\bf bolded}. Epochs that are consumed by the initialization model are \demph{de-emphasized}.  }
    \centering
    \begin{tabular}{m{2.7cm}|m{1.75cm}|m{1.5cm}<{\centering}|m{1.6cm}<{\centering}m{1.75cm}<{\centering}m{1.6cm}<{\centering}}
    method & backbone & epoch & PASCAL & Cityscapes & ADE20k\\
    \Xhline{3\arrayrulewidth}
    supervised & ResNet-50 & - & 73.7 & 75.8 & 37.4\\
    \hline
    MoCo v2~\cite{mocov2} & ResNet-50 & 200 & 74.4 & 75.8 & 37.4\\
    \rowcolor{gray!20}
    {\bf CP$^2$} & ResNet-50 & 200 & 75.4 & 76.4 & 38.4\\\rowcolor{gray!20}
    {\bf CP$^2$} QT r.200 & ResNet-50 & \demph{200+}20 & 75.2 & 76.4 & 38.0\\
    MoCo v2~\cite{mocov2} & ResNet-50 & 800 & 74.8 & 75.9 & 37.9\\
    \rowcolor{gray!20}
    {\bf CP$^2$} QT r.800 & ResNet-50 & \demph{800+}20 & {\bf 75.7} & {\bf 76.5} & {\bf 39.2}\\
    \Xhline{2\arrayrulewidth}
    MoCo v2~\cite{mocov2} & ViT-S/16 & 300 & 77.7 & 76.6 & 40.4\\
    \rowcolor{gray!20}
    {\bf CP$^2$} QT v.300 & ViT-S/16 & \demph{300+}20 & {\bf 78.6} & {\bf 77.0} & {\bf 41.2}\\
    \end{tabular}
    \label{tab:main2}
\end{table}

\textbf{Segmentation head initialization.} Intuitively, CP$^2$ can benefit the downstream segmentation tasks in two aspects. First, CP$^2$ provides the downstream semantic segmentation with a well-pretrained decoder head. Second, CP$^2$ pretrains the model with a segmentation-oriented objective (the dense contrastive loss), which is expected to enable the backbone to extract pixel-level features. To ablate the benefit of each component, we dissect the CP$^2$ trained model and examine the benefits of its backbone and segmentation head respectively.

Table~\ref{tab:abl1} summarizes the results. By pretraining the ResNet-50 based model for 200 epochs from scratch and finetuning on PASCAL, CP$^2$ achieves 77.6\% mIoU, which is 2.7\% higher than that of its MoCo v2 baseline. If we use the CP$^2$ pretrained backbone but still randomly initialize the segmentation head in the finetuning stage, it also attains 1.4\% points higher mIoU than MoCo v2, demonstrating that the backbone representation is also improved for downstream segmentation thanks to our segmentation-oriented objective.

Similarly, the same phenomenon is observed for CP$^2$ Quick Tuning protocol as well. For example, by Quick Tuning the MoCo v2 800-epoch ResNet-50 and finetuning on PASCAL, we obtain +1.4\% mIoU over MoCo v2. While finetuning the CP$^2$ pretrained backbone with a randomly initialized segmentation head also yields +1.0\% mIoU over its baseline.

According to the observation above, CP$^2$ yields both a better backbone and a pretrained segmentation head for downstream semantic segmentation, thanks to our design of CP$^2$ that enables segmentation head pretraining and employs the segmentation-oriented contrastive objective.

\begin{table}[t]
    \caption{\textbf{Ablation study of segmentation head pretraining} on PASCAL VOC. The results are based on ASPP segmentation head. We use Quick Tuning for CP$^2$ in the settings of (ResNet-50, 800 epochs) and (ViT-S/16, 300 epochs).}
    \centering
    \begin{tabular}{m{4cm}|m{2cm}|m{2cm}|m{2cm}}
    mode & backbone & head & mIoU\\\Xhline{3\arrayrulewidth}
    \multirow{3}{*}{ResNet-50, 200 epochs} & MoCo v2 & random & 74.9\\
    & {\bf CP$^2$} & random & 76.3 (+1.4) \\
    & {\bf CP$^2$} & {\bf CP$^2$} & \textbf{77.6} (+\textbf{2.7}) \\\hline
    \multirow{3}{*}{ResNet-50, 800 epochs} & MoCo v2 & random & 77.2\\
    & {\bf CP$^2$} QT & random & 78.2 (+1.0) \\
    & {\bf CP$^2$} QT & {\bf CP$^2$} QT & \textbf{78.6} (+\textbf{1.4}) \\\Xhline{2\arrayrulewidth}
    \multirow{3}{*}{ViT-S/16, 300 epochs} & MoCo v2 & random & 78.8\\
    & {\bf CP$^2$} QT & random & 79.3 (+0.5) \\
    & {\bf CP$^2$} QT & {\bf CP$^2$} QT & \textbf{79.5} (+\textbf{0.7}) \\
    \end{tabular}
    \label{tab:abl1}
\end{table}

\textbf{Foreground-background mask.} We further explore the effect of various types of copy-paste masking for CP$^2$. The experiments are conducted using our Quick Transfer protocol, initialized by the 800-epoch MoCo v2 checkpoint when using a ResNet-50 backbone and the 300-epoch MoCo v2 checkpoint when using a ViT-S backbone, on PASCAL VOC dataset.

First, we consider a baseline when copy-paste masking is not applied. Specifically, the augmented views of the foreground image will \textit{not} be composed with random backgrounds but serve as the model inputs directly. And the model will be trained with only the image-wise contrastive loss ($\mathcal{L}_{ins}$) since there is no background to construct the dense contrastive loss. In other words, the segmentation model (a backbone followed by segmentation head) is simply trained with a MoCo loss that operates on average pooled features over the whole image. We denote this setup as \textit{no copy-paste}. As shown in Table~\ref{tab:mask}, \textit{no copy-paste} yields relatively poor performance. Compared to the baseline performance of 77.2\% mIoU with ResNet-50 and 78.8\% mIoU with ViT-S (Table~\ref{tab:main1}), \textit{no copy-paste} training attains only marginal improvements of 0.4\% and 0.1\% mIoU respectively. This result indicates that pretraining the segmentation head with classification-oriented objectives cannot yield significant improvements on the downstream performance for semantic segmentation. This suggests that the improvement of CP$^2$ mainly comes from the copy-paste training and the dense contrastive loss.

\begin{figure}[t]
  \centering
  \includegraphics[width=0.9\linewidth]{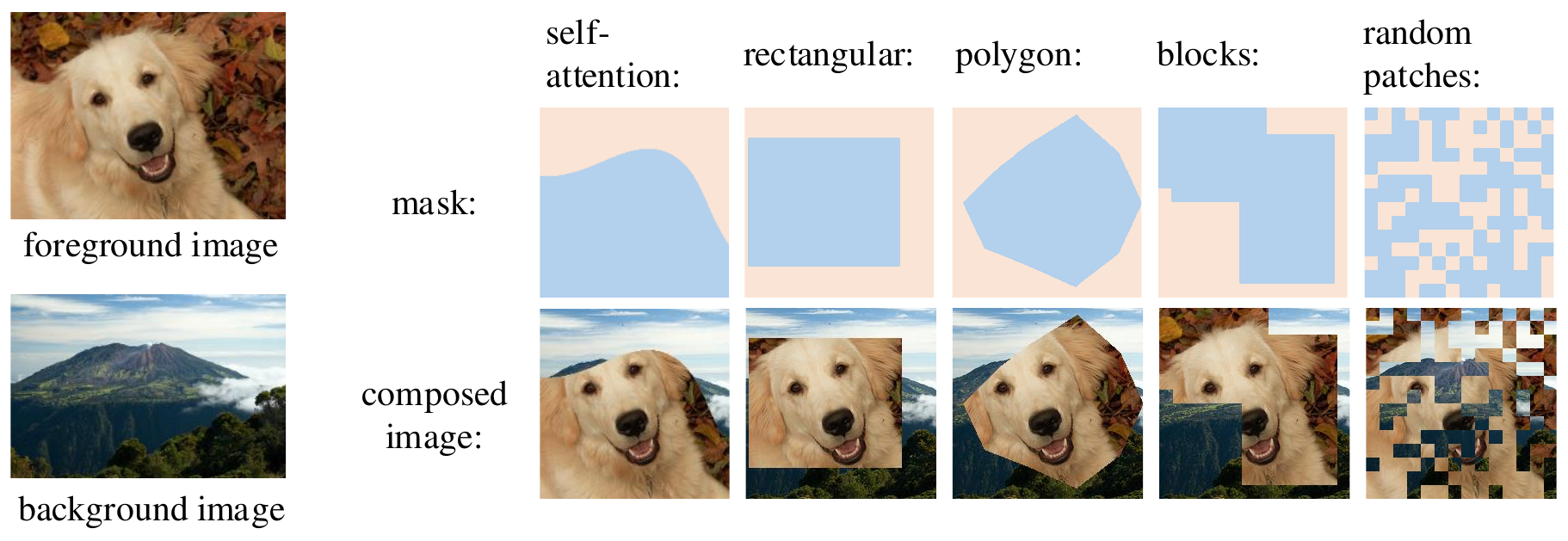}
  \caption{\textbf{Examples of masking strategies and composed images}. The self-attention mask (DINO mask) is smoothed by Gaussian blur.}
  \label{fig:mask}
\end{figure}

We also explore various types of image masking, including the self-attention masks generated by DINO~\cite{dino} and different shapes of random masking. The random masks include rectangular masks, polygon masks, random blocks, and random patches, for which we provide examples in Figure~\ref{fig:mask}. In order to ablate the influence of mask area, we limit the foreground ratio of each random mask to 0.5$\sim$0.8, which we find usually yields better empirical results. The self-attention masks are generated by the DINO~\cite{dino} pretrained ViT-B/16 model. Specifically, for each image, we average their 12 heads of last layer self-attentions and then up-sample the averaged attention map to the original shape of the image. For denoising purpose, we also apply Gaussian blur to the self-attention DINO masks.

\begin{table}[t]
    \caption{\textbf{Evaluation results of foreground-background masks} on PASCAL VOC. Note that for the full mask, the models are trained without dense contrastive loss. Our default setting is marked in \colorbox{gray!20}{gray}.}
    \centering
    \begin{tabular}{m{4.5cm}|m{1.25cm}<{\centering}|m{0.2cm}m{1.5cm}<{\centering}m{1.5cm}<{\centering}}
    \multirow{2}{*}{mode} & \multirow{2}{*}{random}& &  \multicolumn{2}{c}{mIoU}\\
    \cline{4-5}
    &&& ResNet-50 & ViT-S/16\\
    \Xhline{3\arrayrulewidth}
    baseline MoCo v2 & - & & 77.2 & 78.8 \\ \hline
    no copy-paste & - & & 77.6 & 78.9\\
    DINO self-attention mask~\cite{dino} & \xmark & & 77.9 & 79.3\\
    \rowcolor{gray!20}
    rectangular mask & \cmark & & {\bf 78.6} & {\bf 79.5}\\
    polygon mask & \cmark & & 78.1 & 79.0\\
    random blocks & \cmark & & 77.3 & 78.7 \\
    random patches & \cmark & & 75.3 & 78.9\\
    \end{tabular}
    \label{tab:mask}
\end{table}
Empirically, the random rectangular mask achieves the highest performance with both ResNet-50 and ViT-S/16 models in Table~\ref{tab:mask}. This is possibly because the rectangular masks 
contain mostly the real continuous foreground of an image (compared with random patch masks) and also introduce randomness in the masks (compared with DINO masks). This result indicates that simple foreground-background information is sufficient for the models to learn semantic features, and applying the random rectangular masks yields consistent performance gain with both of the backbone architectures. Therefore, we use this simple and easy-to-implement masking strategy in CP$^2$ for the best performance.

Moreover, the ViT-backed model performs more robustly to various shapes of the mask than the ResNet-backed model. It is worth-noting that the {\bf random patches} mask appears to mislead the ResNet-backed model as it yields 75.3\% mIoU, which is 3.3\% lower than the result of rectangular mask and even 1.9\% lower than the MoCo v2 baseline before Quick Tuning. But the ViT model is robust to this random patch masking although no improvement is observed.

\begin{table}[t]
\centering
\caption{\textbf{Evaluation results of hyper-parameter search} on PASCAL VOC. The results are based on ResNet50-ASPP models, where the base backbone is loaded from the MoCo v2 pretrained ResNet50 for 800 epochs. Our default setting is marked in \colorbox{gray!20}{gray}. The best results are \textbf{bolded}.}
\subfloat[\textbf{loss weight and temperature}\label{subtab:param}]{
  \begin{tabular}{m{1.25cm}<{\centering}|m{1.25cm}<{\centering}|m{1.25cm}<{\centering}|m{1.25cm}<{\centering}|m{1.25cm}<{\centering}}
    & \multicolumn{4}{c}{temperature($\tau_{dense}$)}\\
    \cline{2-5}
    weight & 2 & 1 & 0.5 & 0.2\\
    \Xhline{3\arrayrulewidth}
    10 & 77.4 & 77.0 & 76.9 & 77.2\\
    1 & 77.3 & 77.9 & 77.3 & 77.4\\
    0.5 & 77.2 & 78.0 & 77.3 & 77.1\\
    0.2 & 76.9 & \colorbox{gray!20}{\textbf{78.6}} & 77.3 & 76.7\\
    0.1 & 76.0 & 77.7 & 77.5 & 75.8\\
    \end{tabular}
}
\subfloat[\textbf{Quick Tuning epochs}\label{subtab:epoch}]{
  \begin{tabular}{m{1.25cm}<{\centering}|m{2.5cm}<{\centering}}
    epoch & mIoU \\\Xhline{3\arrayrulewidth}
    0 & 77.2 \\
    10 & 77.7 (+0.5) \\\rowcolor{gray!20}
    20 & \textbf{78.6 (+1.4)} \\
    40 & 78.7 (+1.5)\\
    \end{tabular}
}
\label{tab:img-hog}
\end{table}

\textbf{Hyper-parameter search.} It is important to consider the trade-off between the image-wise and pixel-wise objective. Two hyper-parameters, the weight and temperature ($\tau_{dense}$) of pixel-wise contrastive loss (setting the weight of image-wise loss to 1), play decisive roles influencing this trade-off. We conduct grid search of these two parameters and summarize the results in Table~\ref{subtab:param}. As reported, the parameter pair we use in the main experiments, (weight=0.2, $\tau_{dense}$=1), achieves the peak performance. For better efficiency, we recommend a training time of 20 epochs on ImageNet when using Quick Tuning. As listed in Table~\ref{subtab:epoch}, 20 epochs of Quick Tuning yields 78.6\% mIoU (1.4\% higher than MoCo v2) while the 40-epoch Quick Tuning brings only 0.1\% extra improvement.

\section{Conclusion}
\label{sec:concl}

In this work, we propose a segmentation-oriented contrastive learning method CP$^2$, in which we encourage the model to learn both image-level and pixel-level representation by pretraining it with both instance and dense contrastive losses. We point out two key merits of CP$^2$: First, CP$^2$ trains the entire semantic segmentation model, pretraining both the backbone and decoder head, which directly addresses the issue of architectural misalignment when finetuning in downstream semantic segmentation. Second, CP$^2$ is trained on copy-pasted images (images with foreground and background) with a pixel-level dense objective, which helps the model learn localized or spatially varying features that benefit the downstream segmentation task. Our results demonstrate a significant margin over existing methods on semantic segmentation.

\section*{Acknowledgements}
This work was supported by ONR N00014-21-1-2812, NSFC 62176159, Natural Science Foundation of Shanghai 21ZR1432200 and Shanghai Municipal Science and Technology Major Project 2021SHZDZX0102.

\bibliographystyle{splncs04}
\bibliography{ref}

\begin{thebibliography}{10}
\providecommand{\url}[1]{\texttt{#1}}
\providecommand{\urlprefix}{URL }
\providecommand{\doi}[1]{https://doi.org/#1}

\bibitem{arbelaez2014multiscale}
Arbel{\'a}ez, P., Pont-Tuset, J., Barron, J.T., Marques, F., Malik, J.:
  Multiscale combinatorial grouping. In: CVPR (2014)

\bibitem{amdim}
Bachman, P., Hjelm, R.D., Buchwalter, W.: Learning representations by
  maximizing mutual information across views. In: NeurIPS (2019)

\bibitem{belghazi2019learning}
Belghazi, M., Oquab, M., Lopez-Paz, D.: Learning about an exponential amount of
  conditional distributions. In: NeurIPS (2019)

\bibitem{deepcluster}
Caron, M., Bojanowski, P., Joulin, A., Douze, M.: Deep clustering for
  unsupervised learning of visual features. In: ECCV (2018)

\bibitem{swav}
Caron, M., Misra, I., Mairal, J., Goyal, P., Bojanowski, P., Joulin, A.:
  Unsupervised learning of visual features by contrasting cluster assignments.
  In: NeurIPS (2020)

\bibitem{dino}
Caron, M., Touvron, H., Misra, I., J\'egou, H., Mairal, J., Bojanowski, P.,
  Joulin, A.: Emerging properties in self-supervised vision transformers. In:
  ICCV (2021)

\bibitem{chen2018deeplabv2}
Chen, L.C., Papandreou, G., Kokkinos, I., Murphy, K., Yuille, A.L.: Deeplab:
  Semantic image segmentation with deep convolutional nets, atrous convolution,
  and fully connected crfs. IEEE TPAMI  (2017)

\bibitem{chen2017deeplabv3}
Chen, L.C., Papandreou, G., Schroff, F., Adam, H.: Rethinking atrous
  convolution for semantic image segmentation. In: CVPR (2017)

\bibitem{deeplabv3plus2018}
Chen, L.C., Zhu, Y., Papandreou, G., Schroff, F., Adam, H.: Encoder-decoder
  with atrous separable convolution for semantic image segmentation. In: ECCV
  (2018)

\bibitem{simclr}
Chen, T., Kornblith, S., Norouzi, M., Hinton, G.: A simple framework for
  contrastive learning of visual representations. In: ICML (2020)

\bibitem{simclrv2}
Chen, T., Kornblith, S., Swersky, K., Norouzi, M., Hinton, G.: Big
  self-supervised models are strong semi-supervised learners. In: NeurIPS
  (2020)

\bibitem{mocov2}
Chen, X., Fan, H., Girshick, R., He, K.: Improved baselines with momentum
  contrastive learning. arXiv preprint arXiv:2003.04297  (2020)

\bibitem{simsiam}
Chen, X., He, K.: Exploring simple siamese representation learning. In: CVPR
  (2021)

\bibitem{mocov3}
Chen, X., Xie, S., He, K.: An empirical study of training self-supervised
  vision transformers. In: ICCV (2021)

\bibitem{mmseg2020}
Contributors, M.: {MMSegmentation}: Openmmlab semantic segmentation toolbox and
  benchmark. \url{https://github.com/open-mmlab/mmsegmentation} (2020)

\bibitem{cordts2016cityscapes}
Cordts, M., Omran, M., Ramos, S., Rehfeld, T., Enzweiler, M., Benenson, R.,
  Franke, U., Roth, S., Schiele, B.: The cityscapes dataset for semantic urban
  scene understanding. In: CVPR (2016)

\bibitem{imagenet}
Deng, J., Dong, W., Socher, R., Li, L.J., Li, K., Fei-Fei, L.: {ImageNet}: A
  large-scale hierarchical image database. In: CVPR (2009)

\bibitem{pos}
Doersch, C., Gupta, A., Efros, A.A.: Unsupervised visual representation
  learning by context prediction. In: ICCV (2015)

\bibitem{bigan}
Donahue, J., Kr{\"a}henb{\"u}hl, P., Darrell, T.: Adversarial feature learning.
  In: ICLR (2016)

\bibitem{bigbigan}
Donahue, J., Simonyan, K.: Large scale adversarial representation learning. In:
  NeurIPS (2019)

\bibitem{vit}
Dosovitskiy, A., Beyer, L., Kolesnikov, A., Weissenborn, D., Zhai, X.,
  Unterthiner, T., Dehghani, M., Minderer, M., Heigold, G., Gelly, S.,
  Uszkoreit, J., Houlsby, N.: An image is worth 16x16 words: Transformers for
  image recognition at scale. In: ICLR (2021)

\bibitem{dumoulin2016adversarially}
Dumoulin, V., Belghazi, I., Poole, B., Mastropietro, O., Lamb, A., Arjovsky,
  M., Courville, A.: Adversarially learned inference. In: ICLR (2016)

\bibitem{pascal}
Everingham, M., Eslami, S.A., Van~Gool, L., Williams, C.K., Winn, J.,
  Zisserman, A.: The pascal visual object classes challenge: A retrospective.
  IJCV  (2015)

\bibitem{felzenszwalb2004efficient}
Felzenszwalb, P.F., Huttenlocher, D.P.: Efficient graph-based image
  segmentation. IJCV  (2004)

\bibitem{ghiasi2021simple}
Ghiasi, G., Cui, Y., Srinivas, A., Qian, R., Lin, T.Y., Cubuk, E.D., Le, Q.V.,
  Zoph, B.: Simple copy-paste is a strong data augmentation method for instance
  segmentation. In: CVPR (2021)

\bibitem{byol}
Grill, J.B., Strub, F., Altch\'{e}, F., Tallec, C., Richemond, P., Buchatskaya,
  E., Doersch, C., Avila~Pires, B., Guo, Z., Gheshlaghi~Azar, M., Piot, B.,
  kavukcuoglu, k., Munos, R., Valko, M.: Bootstrap your own latent a new
  approach to self-supervised learning. In: NeurIPS (2020)

\bibitem{vocaug}
Hariharan, B., Arbel{\'a}ez, P., Bourdev, L., Maji, S., Malik, J.: Semantic
  contours from inverse detectors. In: ICCV (2011)

\bibitem{moco}
He, K., Fan, H., Wu, Y., Xie, S., Girshick, R.: Momentum contrast for
  unsupervised visual representation learning. In: CVPR (2020)

\bibitem{resnet}
He, K., Zhang, X., Ren, S., Sun, J.: Deep residual learning for image
  recognition. In: CVPR (2016)

\bibitem{henaff2021efficient}
H{\'e}naff, O.J., Koppula, S., Alayrac, J.B., Oord, A.v.d., Vinyals, O.,
  Carreira, J.: {Efficient Visual Pretraining with Contrastive Detection}. In:
  ICCV (2021)

\bibitem{augmix}
Hendrycks, D., Mu, N., Cubuk, E.D., Zoph, B., Gilmer, J., Lakshminarayanan, B.:
  Augmix: A simple data processing method to improve robustness and
  uncertainty. In: ICLR (2019)

\bibitem{supervisedcl}
Khosla, P., Teterwak, P., Wang, C., Sarna, A., Tian, Y., Isola, P., Maschinot,
  A., Liu, C., Krishnan, D.: Supervised contrastive learning. In: NeurIPS
  (2020)

\bibitem{rotnet}
Komodakis, N., Gidaris, S.: Unsupervised representation learning by predicting
  image rotations. In: ICLR (2018)

\bibitem{long2015fully}
Long, J., Shelhamer, E., Darrell, T.: Fully convolutional networks for semantic
  segmentation. In: CVPR (2015)

\bibitem{adamw}
Loshchilov, I., Hutter, F.: Decoupled weight decay regularization. In: ICLR
  (2019)

\bibitem{jigsaw}
Noroozi, M., Favaro, P.: Unsupervised learning of visual representations by
  solving jigsaw puzzles. In: ECCV (2016)

\bibitem{o2020unsupervised}
O~Pinheiro, P.O., Almahairi, A., Benmalek, R., Golemo, F., Courville, A.C.:
  Unsupervised learning of dense visual representations. In: NIPS (2020)

\bibitem{oord2018representation}
Oord, A.v.d., Li, Y., Vinyals, O.: Representation learning with contrastive
  predictive coding. arXiv preprint arXiv:1807.03748  (2018)

\bibitem{pathak2016context}
Pathak, D., Krahenbuhl, P., Donahue, J., Darrell, T., Efros, A.A.: Context
  encoders: Feature learning by inpainting. In: CVPR (2016)

\bibitem{szegedy2017inception}
Szegedy, C., Ioffe, S., Vanhoucke, V., Alemi, A.A.: Inception-v4,
  inception-resnet and the impact of residual connections on learning. In: AAAI
  (2017)

\bibitem{infomin}
Tian, Y., Sun, C., Poole, B., Krishnan, D., Schmid, C., Isola, P.: What makes
  for good views for contrastive learning? In: NeurIPS (2020)

\bibitem{wei2020co2}
Wei, C., Wang, H., Shen, W., Yuille, A.: {CO2}: Consistent contrast for
  unsupervised visual representation learning. In: ICLR (2021)

\bibitem{iterjigsaw}
Wei, C., Xie, L., Ren, X., Xia, Y., Su, C., Liu, J., Tian, Q., Yuille, A.L.:
  Iterative reorganization with weak spatial constraints: Solving arbitrary
  jigsaw puzzles for unsupervised representation learning. In: CVPR (2019)

\bibitem{wu2018unsupervised}
Wu, Z., Xiong, Y., Yu, S.X., Lin, D.: Unsupervised feature learning via
  non-parametric instance discrimination. In: CVPR (2018)

\bibitem{xie2021propagate}
Xie, Z., Lin, Y., Zhang, Z., Cao, Y., Lin, S., Hu, H.: Propagate yourself:
  Exploring pixel-level consistency for unsupervised visual representation
  learning. In: CVPR (2021)

\bibitem{yang2021instance}
Yang, C., Wu, Z., Zhou, B., Lin, S.: Instance localization for self-supervised
  detection pretraining. In: CVPR (2021)

\bibitem{ye2019unsupervised}
Ye, M., Zhang, X., Yuen, P.C., Chang, S.F.: Unsupervised embedding learning via
  invariant and spreading instance feature. In: CVPR (2019)

\bibitem{cutmix}
Yun, S., Han, D., Oh, S.J., Chun, S., Choe, J., Yoo, Y.: Cutmix: Regularization
  strategy to train strong classifiers with localizable features. In: ICCV
  (2019)

\bibitem{mixup}
Zhang, H., Cisse, M., Dauphin, Y.N., Lopez-Paz, D.: mixup: Beyond empirical
  risk minimization. In: ICLR (2017)

\bibitem{zhang2016colorful}
Zhang, R., Isola, P., Efros, A.A.: Colorful image colorization. In: ECCV (2016)

\bibitem{zhang2017split}
Zhang, R., Isola, P., Efros, A.A.: Split-brain autoencoders: Unsupervised
  learning by cross-channel prediction. In: CVPR (2017)

\bibitem{zhao2021contrastive}
Zhao, X., Vemulapalli, R., Mansfield, P.A., Gong, B., Green, B., Shapira, L.,
  Wu, Y.: Contrastive learning for label efficient semantic segmentation. In:
  ICCV (2021)

\bibitem{zhou2019semantic}
Zhou, B., Zhao, H., Puig, X., Xiao, T., Fidler, S., Barriuso, A., Torralba, A.:
  Semantic understanding of scenes through the ade20k dataset. IJCV  (2019)

\bibitem{zhou2022ibot}
Zhou, J., Wei, C., Wang, H., Shen, W., Xie, C., Yuille, A., Kong, T.: {iBOT}:
  Image {BERT} pre-training with online tokenizer. In: ICLR (2022)

\end{thebibliography}

\section*{Appendix}

\subsubsection{Additional ablations.} We have observed that training CP$^2$ without copy-paste and dense contrastive loss yields poor downstream performance. To further ablate this effect, we pretrain CP$^2$ models with copy-pasted images but using only one loss term, \ie the dense contrastive loss $\mathcal{L}_{dense}$ or the instance contrastive loss $\mathcal{L}_{ins}$. This ablation study further dissects the components of CP$^2$ and gives better comprehension of their effects. For better efficiency, we conduct this ablation on top of the MoCo v2 800-epoch ResNet-50 and apply CP$^2$ Quick Tuning for 20 epochs. As listed in Table~\ref{tab:apdx1}, compared with the MoCo v2 baseline, applying Quick Tuning protocol in the MoCo manner (no copy-paste) slightly improves the downstream performance by 0.4\% mIoU, while employing the entire CP$^2$ design yields 1.4\% mIoU improvements.

\begin{table}[htbp]
\vspace{-10pt}
    \caption{{\bf Ablation results} of CP$^2$. The results (mIoU on PASCAL VOC) are based on ResNet50-ASPP models, where the base backbone is loaded from the MoCo v2 pretrained ResNet50 for 800 epochs.}
    \centering
    \begin{tabular}{m{2.5cm}|m{1cm}<{\centering}m{1.75cm}<{\centering}m{1.5cm}<{\centering}m{2cm}<{\centering}|m{2cm}<{\centering}}
    mode & QT & copy-paste & dense loss & instance loss & mIoU\\\Xhline{3\arrayrulewidth}
    baseline & & {\bf -} & {\bf -} & {\bf -} & 77.2\\
    no copy-paste & \cmark & & & \cmark & 77.6\\
    instance loss only & \cmark & \cmark & & \cmark & 78.0\\
    dense loss only & \cmark & \cmark & \cmark & & 75.6\\
    entire CP$^2$ & \cmark & \cmark & \cmark & \cmark & 78.6\\
    \end{tabular}
    \label{tab:apdx1}
\vspace{-10pt}
\end{table}

We also note that despite omitting the dense contrastive loss of CP$^2$ (instance loss only) leads to 0.6\% mIoU degradation (compared with entire CP$^2$), it outperforms Quick Tuning with no copy-pasted images. The only difference between the setup of {\it no copy-paste} and {\it instance loss only} is that the former uses original augmented views and apply {\bf global average pooling} to the dense features for instance discrimination, while the later uses copy-pasted views and {\bf masked pooling}. Therefore, this result indicates that training with copy-pasted images provides the models with better robustness to background noise, which benefits the downstream performance of semantic segmentation. Moreover, omitting the instance contrastive loss of CP$^2$ yields even worse downstream performance (dense loss only) than the MoCo v2 baseline, which indicates that the capability of instance discrimination is one of the models' key components for semantic segmentation.

\begin{table}[htbp]
\vspace{-10pt}
    \caption{{\bf Deviations of CP$^2$ results}. The results are based on ResNet50-ASPP models, where the base backbone is loaded from the MoCo v2 pretrained ResNet50 for 800 epochs. We run three trials for mean$\pm$std results.}
    \centering
    \begin{tabular}{m{2cm}<{\centering}|m{2cm}<{\centering}m{2cm}<{\centering}m{2cm}<{\centering}}
    dataset & PASCAL & Cityscapes & ADE20k \\\Xhline{3\arrayrulewidth}
    mIoU & 78.6$\pm$0.2 & 77.4$\pm$0.2 & 41.3$\pm$0.1
    \end{tabular}
    \label{tab:std}
\vspace{-10pt}
\end{table}

\subsubsection{Deviations of main results.} To demonstrate the robustness of CP$^2$, here we in addition report the standard deviations of our main experimental results. We run three trials of CP$^2$ Quick Tuning and downstream finetuning and list the mean$\pm$std results in Table~\ref{tab:std}.

\end{document}